\documentclass{article}
\usepackage{booktabs,amsthm}

\usepackage{amssymb}

\usepackage{relsize}
\usepackage{framed,amsmath}
\usepackage{algorithm2e}
\usepackage{tikz}
\usetikzlibrary{shapes.geometric,positioning}
\usepackage{pgfplots}
\usepackage{amsthm}

\usepackage{tikz}
\usepackage{subfig}
\usepackage{pgfplots}
\usepackage{manyfoot}
\usepackage{multirow}
\usepackage{graphicx}

\begin{document}

\title{Custom Extended Sobel Filters\\
(extended abstract)}  

\author{Victor BOGDAN$^1$, Cosmin BONCHI\c{S}$^1$, Ciprian ORHEI$^2$\\
$1$West University of Timi\c{s}oara and the eAustria Research Institute, \\Bd. V. P\^arvan 4, cam 045B, Timi\c{s}oara, RO-300223, Romania.\\
$2$Politehnica University of Timisoara, Timi\c{s}oara, RO-300223, Romania.}

\maketitle

\abstract{Edge detection is widely and fundamental feature used in various algorithms in computer vision to determine the edges in an image. The edge detection algorithm is used to determine the edges in an image which are further used by various algorithms from line detection to machine learning that can determine objects based on their contour. Inspired by new convolution techniques in machine learning we discuss here the idea of extending the standard Sobel kernels, which are used to compute the gradient of an image in order to find its edges. We compare the result of our custom extended filters with the results of the standard Sobel filter and other edge detection filters using different image sets and algorithms. We present statistical results regarding the custom extended Sobel filters improvements. 

}

\section{Introduction}
The Sobel filter is used in many algorithms which rely on edge detection for a range of applications: e.g. license plates recognition of cars and successfully retrieve it in the format of a string. \cite{reddy2017efficient} applies Sobel filter on a pre-processed image in order to retrieve an edge image, used to find and extract a rectangular area in the original image which represents the license plate. The Sobel filter was used also in criminal forensics and personal identification \cite{bandyopadhyay2013feature}. It was used together with the Canny Edge Detector to show the vertical and horizontal groove pattern in the lip. The Sobel filter is also used in the pre-processing of blood perfusion images resulted from thermograms of human faces alongside two other methods described in \cite{seal2013automated}. 

Although, the Sobel filter is widely used in edge detection, there are also other methods which prove to be more efficient. The SVM described in \cite{irandoust2017gaussian} successfully identifies edges in images using a Gaussian Three-Dimensional kernel. This algorithm has a better performance compared to the standard Sobel and Canny edge detectors.

In \cite{gupta2013sobel} they have built an extended $5\times5$ Sobel filter and they mathematically proved how to build it starting from classical $3\times3$ Sobel operator. They conclude that a bigger filter is useful in order to find more edge pixels. However, in our approach we use a simplified extended version of the standard Sobel filter, that seems to obtain surprisingly better results in practice. The first motivation to use the simplified version of this extended filter is the runtime efficiency, see the details in Section \ref{more_mathematical_filters_comparisons}.

In this paper we will present the extended Sobel filters used to compute gradient of an image in order to find its edges. We compare the result of our extended filters with the standard and extended  Prewitt, Sobel and Scharr filters, presented in \cite{levkine2012prewitt}, using different image sets. We used the BSDS500 benchmark tool and image sets from \cite{amfm_pami2011} for the comparison results. 

We used the Canny Edge detection algorithm \cite{canny86} in order to compare our extended filter with the standard approach that uses the  classical $3\times3$ Sobel operator.

By replacing the Sobel filter with the extended versions ($5\times5$, $7\times7$ up to $15\times15$) of it, we analyze which of those extended filters can improve the most the Canny Edge detection algorithm in practice.

In the following we present the standard Sobel algorithm and notations in Section \ref{Sec:Preliminaries}, the definition of the extended filters for different sizes are presented in \ref{Sec:Extended filters}, then the Section \ref{Sec:Simulation results} is containing all the comparison simulation and the intuitions of the results.

\section{Preliminaries}
\label{Sec:Preliminaries}
We will consider the following standard formula where the $Gx$ and $Gy$ gradient components are used to define the gradient magnitude $|G|$:
\begin{align}\label{magnitude}
|G| = \sqrt{Gx^2 + Gy^2}.
\end{align} 

We use in the paper the edge detection algorithm steps, presented in \cite{Woods2011}, which are generally used to convolve the filters with a source image in order to obtain the edge image: 

\begin{itemize}
\item[Step 1] Convert the image to gray-scale, preparing as input for Sobel filter convolution.	
\item[Step 2] Reduce the noise in the source image by applying the Gaussian filter in order to obtain smooth continuous values. 
\item[Step 3] Applying the filters by convolving the gray-scale image with their kernels on the $x$ and $y$ axes and then applying the gradient magnitude (Equation \ref{magnitude}).
\item[Step 4] Each pixel which has an intensity value higher or equal to a \textit{threshold} will have its value set to MaxValue (e.g. 255), else to 0, therefore the edges will be represented by the white pixels. 
\end{itemize}

In the comparison results, the applied threshold was chosen accordingly by the benchmarking tool computations from BSDS500. 

We also analyze the results of the Canny edge algorithm using the Sobel filter and the custom extended filters. 
Canny edge detection is a widely known and one of the most used edge detection algorithms \cite{canny86}. 
We compare the effect of the filters on the test, train and validation image sets from BSDS500. 
The algorithm can be broken down to 4 steps, assuming that the input image is a gray-scale image:

\begin{enumerate}
  \item Applying the Gaussian Filter.
  \item Finding the magnitude and orientation of the gradient.
  \item Non-maximum suppression.
  \item Edge tracking by hysteresis using double threshold. 
\end{enumerate}

The double threshold is found, by using the maximum pixel intensity in the input image and applying the formula from Equation \ref{ratio_threshold} where $\mathit{T}_{h}$ represents the upper threshold, $\mathit{T}_{l}$ the lower threshold and $max(input)$ is the maximum pixel intensity in the input image:
\begin{align}
\label{ratio_threshold}
&\mathit{T}_{h} = max(input)\times0.7 \\
&\mathit{T}_{l} = \mathit{T}_{h}\times0.3 \nonumber
\end{align}

\section{Extended filters}
\label{Sec:Extended filters}
One of the commonly used methods in detecting the edges in an image is by convolving the initial image with the Sobel filter. The filter calculates the difference among the pixel intensities. 
To obtain the edge image the Sobel Operator kernels are applied and combined in order to obtain the higher changes in intensity, by using the Sobel algorithm steps presented in the previous section.

We propose in this paper to use some modified Sobel filters used for detecting the edges in images from a different perspective. The standard $3\times3$ Sobel kernels are extended to $5\times5$, $7\times7$, $9\times9$, $11\times11$, $13\times13$ and $15\times15$ kernels, for both axes. By extending the kernels we propose to increase the distance between the pixels which influence the result of the convolution. This expansion induces the possibility of finding stronger intensity changes in the image. When we extend the kernels, we are filling the newly added kernel positions with 0s, see the Figures \ref{55Sobel0} or \ref{77Sobel0}. With our approach we obtain in most of the test cases better edge detection results than the filters presented in \cite{gupta2013sobel} or \cite{levkine2012prewitt}, even if it doesn't respect the geometrical gradient formulas. Another advantage of our filter is the reduced runtime complexity due to the simplicity of the extension.

\begin{figure}
\small{
\begin{align*}
  \renewcommand{\arraystretch}{1.5}
  \begin{bmatrix}
  1&0&0&0&-1\\
  0&0&0&0&0\\
  2&0&0&0&-2\\
  0&0&0&0&0\\
  1&0&0&0&-1\\
  \end{bmatrix}
  &
  \renewcommand{\arraystretch}{1.5}
  \begin{bmatrix}
  1&0&2&0&1\\
  0&0&0&0&0\\
  0&0&0&0&0\\
  0&0&0&0&0\\
  -1&0&-2&0&-1\\
  \end{bmatrix}
\end{align*}
}
\caption{\label{55Sobel0}Custom $5\times5$ extended Gx and Gy kernels with 0s}
\end{figure}


\begin{figure}
\tiny{
\begin{align*}
  \renewcommand{\arraystretch}{1.5}
  \begin{bmatrix}
  1&0&0&0&0&0&-1\\
  0&0&0&0&0&0&0\\
  0&0&0&0&0&0&0\\
  2&0&0&0&0&0&-2\\
  0&0&0&0&0&0&0\\
  0&0&0&0&0&0&0\\
  1&0&0&0&0&0&-1\\
  \end{bmatrix}
  &
  \renewcommand{\arraystretch}{1.5}
  \begin{bmatrix}
  1&0&0&2&0&0&1\\
  0&0&0&0&0&0&0\\
  0&0&0&0&0&0&0\\
  0&0&0&0&0&0&0\\
  0&0&0&0&0&0&0\\
  0&0&0&0&0&0&0\\
  -1&0&0&-2&0&0&-1\\
  \end{bmatrix}
\end{align*}
}
\caption{\label{77Sobel0} Custom $7\times7$ extended Gx and Gy kernels with 0s}
\end{figure}

\section{Simulation results}
\label{Sec:Simulation results}

\begin{figure}[ht!]\centering
	\begin{minipage}[t]{0.2\linewidth}
	\includegraphics[scale=0.2]{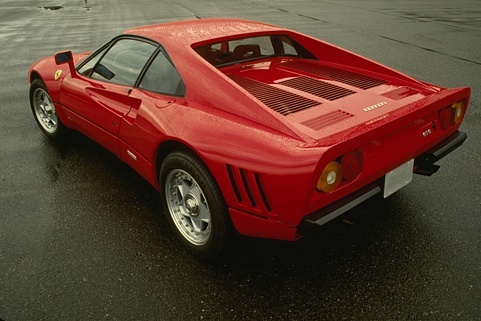}
	\label{fig:subfig1}
	\caption{Original image}
	\end{minipage}
\qquad \qquad \qquad
	\begin{minipage}[t]{0.2\linewidth}
	\includegraphics[scale=0.2]{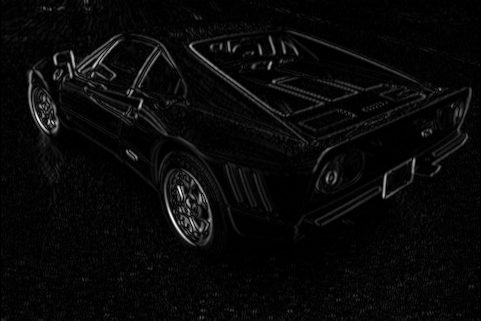}
	\label{fig:subfig2}
	\caption{$3\times3$ standard Sobel}
	\end{minipage}
	
	\begin{minipage}{0.2\linewidth}
	\includegraphics[scale=0.2]{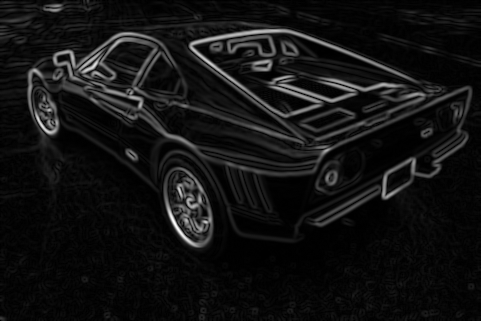}
	\label{fig:subfig3}
	\caption{$5\times5$ extended Sobel}
	\end{minipage}
\qquad \qquad \qquad
	\begin{minipage}{0.2\linewidth}
	\includegraphics[scale=0.2]{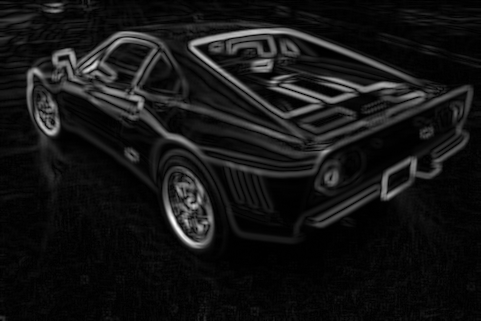}
	\label{fig:subfig4}
	\caption{$7\times7$ extended Sobel}
	\end{minipage}

	\begin{minipage}{0.2\linewidth}	
	\includegraphics[scale=0.2]{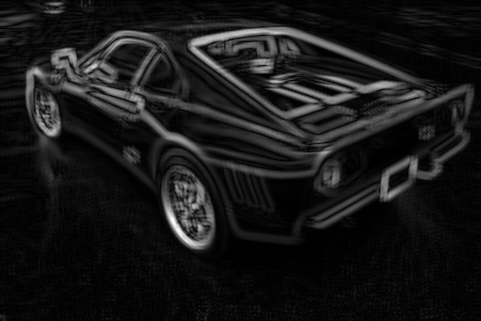}
	\label{fig:subfig5}
	\caption{$9\times9$ extended Sobel}
	\end{minipage}
\qquad \qquad \qquad
	\begin{minipage}{0.2\linewidth}
	\includegraphics[scale=0.2]{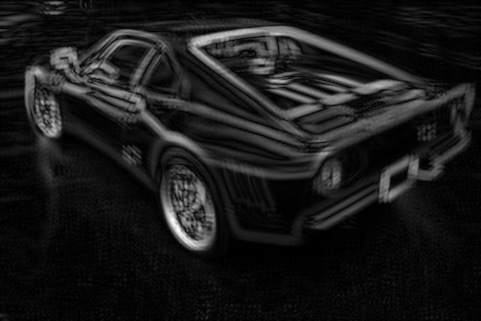}
	\label{fig:subfig6}
	\caption{$11\times11$ extended Sobel}
	\end{minipage}
	
	\begin{minipage}{0.2\linewidth}
	\includegraphics[scale=0.2]{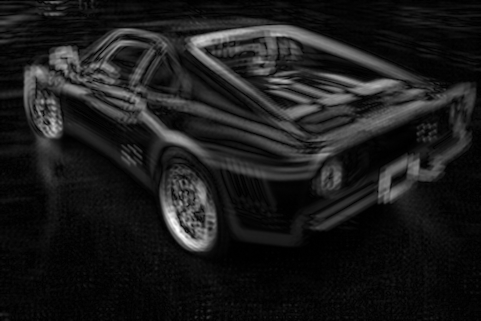}
	\label{fig:subfig7}
	\caption{$13\times13$ extended Sobel}
	\end{minipage}
\qquad \qquad \qquad
	\begin{minipage}{0.2\linewidth}
	\includegraphics[scale=0.2]{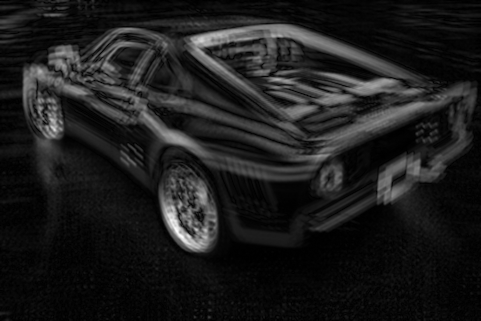}
	\label{fig:subfig8}
	\caption{$15\times15$ extended Sobel}
	\end{minipage}
	
\caption{Edge detection by applying Sobel filters.}
\label{fig:sobel_table}
\end{figure}

In the following section we present the visual comparisons between the results of convolving an image with the standard Sobel filter and the extended filters defined previously. We used the image sets from BSDS500 as a standard edge detection experimental data set in order to compare our extended ($5\times5$, $7\times7$, $\dots$, $15\times15$) filters with the standard Sobel filter, the Figure \ref{fig:sobel_table} contains such visual results. We can see that some extended filters provide better edge detection results than the standard $3\times3$ Sobel filter, that give us the motivation to further analyze the edge detection quality/efficiency of those extensions and search the best answer to the question: Which extension provides the better benchmarking results?


In the resulting edge images (Figure \ref{fig:sobel_table}) we can notice that when we convolve the original source image with the custom extended filters, 
we obtain more pixels with a high gradient magnitude. Also, the more the extension is increased, the more the edge images seem to lose the details while the edges seem to get blurry. There should be a compromise between details and the newly detected edges.
In the following we will use a standard data set and benchmark in order to compare the extended filters.

We used BSDS500 from \cite{amfm_pami2011} as the benchmark tool, that contains 500 test images, which are split in 3 different sets, each having a few human segmented boundary ground-truth images. Our results will be compared to the ground-truth images. The benchmark will choose a number of threshold values between 0 and 255 and thus will determine which is the best possible result for each of the filters for each image in the set.

The threshold for the statistical comparison was determined using the benchmarking tool, which chose from a large set of thresholds, the one that obtains the best overall F1-score, that have the following definition:

The \textit{F-measure (F1-score)} is an accuracy test measure given by: 

\begin{align}\label{f-measure}
F-measure = \frac{(2*TP)}{(2*TP+FP+FN)}.
\end{align}

where we have the true positive (TP), false positive (FP) and false negative (FN) rates. F1-score can also be interpreted as the harmonic mean of precision and sensitivity (recall) standard measures: Precision, Equation \ref{precision}, represents the probability that a resulting edge/boundary pixel is a true edge/boundary pixel. Recall, formula \ref{recall}, represents the probability that a true edge/boundary pixel is detected.

\begin{align}\label{precision}
P = \frac{TP}{(TP+FP)}.
\end{align}

\begin{align}\label{recall}
R = \frac{TP}{(TP+FN)}.
\end{align}

\begin{table}[ht!]
\centering
\caption{\label{tab:sobel_test_set_statistics} Custom extended Sobel Comparison on the Test Set from BSDS500}
\begin{tabular}{ | c | c | c | c | c | c | }
 \hline
 Filter  & Overall 	& Overall 	 & Overall  \\
 Size    & Recall 	& Precision  & F1-score	   \\		
 \hline
  $3\times3$  	& 0.581269	& 0.430275	& 0.494503			\\
  $5\times5$ 	& 0.790220  & 0.426561	& 0.554047 	 		\\
  \boldmath{$7\times7$} &\textbf{0.786260}  &\textbf{0.446317} 	&\textbf{0.569410} 	       \\
  $9\times9$ 	& 0.791705  & 0.439016 	& 0.564825 	 		\\
  $11\times11$ & 0.795737  & 0.426619 	& 0.555446 	 		\\
  $13\times13$ & 0.791030  & 0.416494 	& 0.545677 	 		\\
  $15\times15$ & 0.785332  & 0.406225 	& 0.535470 	 		\\
\hline
\end{tabular}
\vspace{5mm}

\caption{\label{tab:sobel_train_set_statistics} Custom extended Sobel Comparison on the Train Set from BSDS500}
\begin{tabular}{ | c | c | c | c | c | c | }
 \hline
 Filter  & Overall 	& Overall 	 & Overall  \\
 Size    & Recall 	& Precision  & F1-score	   \\		
 \hline
  $3\times3$  	& 0.598837  & 0.402990  & 0.481770			\\
  $5\times5$ 	& 0.806898  & 0.397421  & 0.532548 	 		\\
  \textbf{$7\times7$} 	& \textbf{0.800562}  & \textbf{0.417457}  & \textbf{0.548760} \\
  $9\times9$ 	& 0.496480  & 0.610924  & 0.547788 	 		\\
  $11\times11$ & 0.504355  & 0.601248  & 0.548556 	 		\\
  $13\times13$ & 0.505152  & 0.583671  & 0.541580 	 		\\
  $15\times15$ & 0.502193  & 0.564027  & 0.531317 	 		\\
\hline
\end{tabular}
\vspace{5mm}

\caption{\label{tab:sobel_validation_set_statistics} Custom extended Sobel Comparison on the Validation Set from BSDS500}
\begin{tabular}{ | c | c | c | c | c | c | }
 \hline
 Filter  & Overall 	& Overall 	 & Overall  \\
 Size    & Recall 	& Precision  & F1-score	   \\		
 \hline
  $3\times3$  	& 0.611690  & 0.410208  & 0.491087			\\
  $5\times5$ 	& 0.807394  & 0.403534  & 0.538118 	 		\\
  \textbf{$7\times7$} 	& \textbf{0.800256}  & \textbf{0.423428}  & \textbf{0.553820} \\
  $9\times9$ 	& 0.803704  & 0.415565  & 0.547855 	 		\\
  $11\times11$ & 0.493794  & 0.594842  & 0.539628 	 		\\
  $13\times13$ & 0.501036  & 0.576241  & 0.536014 	 		\\
  $15\times15$ & 0.498039  & 0.557311  & 0.526010 	 		\\
\hline
\end{tabular}
\end{table}

\subsection{Sobel benchmarking results}

We have use a similar setup as in \cite{amfm_pami2011} and the results (overall recall, overall precision and overall F1-score) presented in the following should be interpreted as average scores for all edge image samples. The presented values are obtained for best F1-score from different threshold levels analyzed.

The results of the standard Sobel filter and custom extended Sobel filters on the BSDS500 test set, train set and validation set can be seen in Tables \ref{tab:sobel_test_set_statistics}, \ref{tab:sobel_train_set_statistics} and \ref{tab:sobel_validation_set_statistics}. 

We can observe that every extended Sobel filter obtained a better overall F1-score than the $3\times3$ standard Sobel filter. The best results in all image sets were obtained by the $7\times7$ extended Sobel filter.

By increasing the extension of the Sobel filter, we can also notice that the F1-score starts to decrease, that is a normal behavior due to the false negatives induces by the high gradient distances. The $13\times13$ and $15\times15$ extended Sobel filters obtained a significantly lower F1-score than the $7\times7$ extended filter in all image sets. This is also the case for the overall precision, whereas the overall recall alternates but there are no significant differences. 

An interesting result can be noticed on the train set (Table \ref{tab:sobel_train_set_statistics}) and validation set (Table \ref{tab:sobel_validation_set_statistics}) where the expansion of the filter leads to an increase of precision but decrease of recall, whereas on the test set there was no noticeable difference between the custom extended filters regarding the recall.

\subsection{Extension comparison results}
\label{more_mathematical_filters_comparisons}

In order to validate the extension filter usability we have compared our extended Sobel (Ext.) filter with the mathematical defined extended Sobel (Sob.) filter from \cite{gupta2013sobel} and the extended Prewitt (Prew.), Modified Prewitt (Prew*) and Scharr (Sch.) filters, described in \cite{levkine2012prewitt}. 

Even though we showed previously that $7\times7$ extended filter obtained the best results, we have used for the comparison only the $5\times5$ extended Sobel in order to match with those predefined mathemathical filters, in \cite{gupta2013sobel} and \cite{levkine2012prewitt}.

The results on each set are listed in Table\ref{tab:custom_other_comparison_statistics}, where the $5\times5$ custom extended Sobel filter has obtained a better overall F1-score than all other filters. Some of the other filters could have better recall or precision but at the cost of a lower overall F1-score. The extension simplicity of our filters produces the runtime efficiency that we have observed in our experiments, however the mathematical proof of why we obtained better results than all other filters is still an open point for us. 

\begin{table}[ht!]
\centering
\caption{\label{tab:custom_other_comparison_statistics} Custom $5\times5$ and Sobel $5\times5$ comparisons}
\begin{tabular}{ | c | c | c | c | c | }
 \hline
 \multirow{2}{*}{Set} & Filter  & Overall 	& Overall 	 & Overall  \\
 & Type    & Recall 	& Precision  & F1-score	   \\		
 \hline
  \multirow{5}{*}{\rotatebox{90}{Test Set}} & Sobel $5\times5$  	& 0.795001  & 0.408893  & 0.540032	\\
  & \boldmath{\textbf{Custom} \boldmath{ $5\times5$}} 	& \textbf{0.790220}  & \textbf{0.426561}  & \textbf{0.554047}  \\  
  & Scharr $5\times5$ & 0.795289 & 0.406717 & 0.538196 \\ 
  & Prewitt $5\times5$ & 0.788270 &  0.425267 &  0.552476 \\
  & M.Prewitt $5\times5$ & 0.794205 & 0.411989 & 0.542539\\ \hline
  \multirow{5}{*}{\rotatebox{90}{Train Set}} & Sobel $5\times5$  	& 0.812620  & 0.381503  & 0.519238	\\
  & \boldmath{\textbf{Custom} \boldmath{$5\times5$}} 	& \textbf{0.806898}  & \textbf{0.397421}  & \textbf{0.532548}  \\  
  & Scharr $5\times5$ & 0.813094 & 0.379770 & 0.517726 \\
  & Prewitt $5\times5$ & 0.805370 & 0.396199 & 0.531117 \\
  & M.Prewitt $5\times5$ & 0.811261 & 0.384073 & 0.521333 \\ \hline
  \multirow{5}{*}{\rotatebox{90}{Valid. Set}} & Sobel $5\times5$  	& 0.815561  & 0.385666  & 0.523688	\\
  & \boldmath{\textbf{Custom} \boldmath{$5\times5$}} 	& \textbf{0.807394 }  & \textbf{0.403534}  & \textbf{0.538118}  \\
  & Scharr $5\times5$ & 0.815913 & 0.384072 & 0.522289 \\
  & Prewitt $5\times5$ & 0.807183 & 0.401473 & 0.536235 \\
  & M.Prewitt $5\times5$ & 0.814780 & 0.388225 & 0.525879 \\ \hline
\end{tabular}

\end{table}

\subsection{Canny Edge result comparisons}

The benchmarking results obtained so far do not confirm or infirm that our extension approach works fine embedded in a complex edge detection algorithm. Due to the impressive results obtained in the previous section we compare our extended filters only with the standard $3\times3$ Sobel filter in the Canny edge detection algorithm.

In Tables \ref{tab:canny_test_set_statistics}, \ref{tab:canny_train_set_statistics} and \ref{tab:canny_validation_set_statistics}, we can see the results of the Canny edge algorithm using the standard Sobel filter but also our extended Sobel filters over the test, train and validation sets from BSDS500. We used the filters for computing the magnitude and orientation of the gradient in the Canny edge algorithm. 

The accuracy in all those experiments (see the tables \ref{tab:canny_test_set_statistics}-\ref{tab:canny_validation_set_statistics}) is increasing up to a point ($7\times7$ or $9\times9$ filter sizes) and works highly better that the standard $3\times3$ Sobel filter in the Canny edge detection algorithm in all cases.

\begin{table}[ht!]
\centering

\caption{\label{tab:canny_test_set_statistics} Canny Edge with custom extended Sobel Comparison on the Test Set from BSDS500}
\begin{tabular}{ | c | c | c | c | c | c | }
 \hline
 Filter  & Overall 	& Overall 	 & Overall  \\
 Size    & Recall 	& Precision  & F1-score	   \\		
 \hline
  $3\times3$  	& 0.474017  & 0.548234  & 0.508432			\\
  $5\times5$ 	& 0.599566  & 0.550924  & 0.574217 	 		\\
  \textbf{$7\times7$} 	& \textbf{0.604575}  & \textbf{0.569470}  & \textbf{0.586498} \\
  $9\times9$ 	& 0.600985  & 0.572163  & 0.586220 	 		\\
  $11\times11$ & 0.601120  & 0.562658  & 0.581254 	 		\\
  $13\times13$ & 0.596377  & 0.541971  & 0.567874 	 		\\
  $15\times15$ & 0.590094  & 0.519348  & 0.552465 	 		\\
\hline
\end{tabular}
\vspace{5mm}

\caption{\label{tab:canny_train_set_statistics} Canny Edge with custom extended Sobel Comparison on the Train Set from BSDS500}
\begin{tabular}{ | c | c | c | c | c | c | }
 \hline
 Filter  & Overall 	& Overall 	 & Overall  \\
 Size    & Recall 	& Precision  & F1-score	   \\		
 \hline
  $3\times3$  	& 0.467275  & 0.522583  & 0.493384			\\
  $5\times5$ 	& 0.599827  & 0.526618  & 0.560844 	 		\\
  $7\times7$ 	& 0.600798  & 0.544154  & 0.571075 \\
  \textbf{$9\times9$} 	& \textbf{0.596565}  & \textbf{0.560563}  & \textbf{0.578004} 	 		\\
  $11\times11$ & 0.596875  & 0.550471  & 0.572735 	 		\\
  $13\times13$ & 0.589993  & 0.531686  & 0.559324 	 		\\
  $15\times15$ & 0.583896  & 0.508945  & 0.543851 	 		\\
\hline
\end{tabular}
\vspace{5mm}

\caption{\label{tab:canny_validation_set_statistics} Canny Edge with custom extended Sobel Comparison on the Validation Set from BSDS500}
\begin{tabular}{ | c | c | c | c | c | c | }
 \hline
 Filter  & Overall 	& Overall 	 & Overall  \\
 Size    & Recall 	& Precision  & F1-score	   \\		
 \hline
  $3\times3$  	& 0.452478  & 0.533517  & 0.489667			\\
  $5\times5$ 	& 0.581843  & 0.529641  & 0.554516 	 		\\
  $7\times7$ 	& 0.583648  & 0.552102  & 0.567437 \\
  \textbf{$9\times9$} 	& \textbf{0.578976}  & \textbf{0.557837}  & \textbf{0.568210} 	 		\\
  $11\times11$ & 0.582776  & 0.547655  & 0.564670 	 		\\
  $13\times13$ & 0.579200  & 0.528327  & 0.552595 	 		\\
  $15\times15$ & 0.572947  & 0.506787  & 0.537840 	 		\\
\hline
\end{tabular}

\end{table}

In Table \ref{tab:canny_test_set_statistics} we can notice that the $7\times7$ custom extended filter obtained the best F1-score on the test set from BSDS500. As we extended the filter the F1-score starts to decrease similarly to the previous edge finding technique comparisons. The precision also starts to decrease but there is no significant difference regarding the recall. Tables \ref{tab:canny_train_set_statistics} and \ref{tab:canny_validation_set_statistics} contain similar results for the train and validation image sets, the only difference being that the $9\times9$ custom extended filter obtained the best F1-score by a small extra bound.

\section{Conclusions and future work}

We highlight in this paper that the extending Sobel filter has better results to find more edge pixels than the standard Sobel filter. By extending the kernels of the Sobel filter we obtained a better recall and precision, which can be observed in Tables \ref{tab:sobel_test_set_statistics} - \ref{tab:sobel_validation_set_statistics}, and thus imply a better accuracy F1-score. Visually, the results of the custom extended Sobel filters seem to get more blurry as the size of the extension increases. 

By comparing the custom extended Sobel filter with other edge detection filters, which are built based on a mathematical explanation from \cite{gupta2013sobel} and \cite{levkine2012prewitt}, we can notice in Table \ref{tab:custom_other_comparison_statistics} that the custom extended Sobel $5\times5$ obtained a better overall F1-score than the other filters. 

Because of the simple structure of the custom extended Sobel filters, they are also a good choice when the run-time matters. The other filters from \cite{levkine2012prewitt} require a larger number of operations in order to return the resulting edge pixels, whereas the custom extended Sobel filters have always the same number of operations for any extension.

There are many algorithms which use the Sobel filter to compute the gradient magnitude in order to detect the edges of an image. Canny Edge is one of them and we have used it to compare the extended Sobel filters with the $3\times3$ standard Sobel. In Tables \ref{tab:canny_test_set_statistics} - \ref{tab:canny_validation_set_statistics} we presented the improvements of the recall, precision and F1-score with the Canny Edge algorithm.

Therefore by visually and statistically comparing the custom filters with the standard Sobel filter we can conclude that the custom extended filters can achieve better results than the other known edge detection filters. However, our intuition is that our proposed extensions might not be very efficient in images which contain many details because the extension of the kernels can induce loss of details in images which have high differences of pixel intensities.

Further simulations would possibly give us a way of choosing the best edge detection filter regarding the input image sets and justifying when the extension or the standard filters should be used. Because we saw there is no filter that works better in all the scenarios a further idea is to implement some machine learning techniques which can be used in order to find the best compromise when a filter is more feasible than another for different contexts. 

\section*{\textit{ACKNOWLEDGEMENTS}}

\noindent This work was partially supported by a grant of Ministry of Research and Innovation, CNCS - UEFISCDI, project number
PN-III-P4-ID-PCE-2016-0842, within PNCDI III.

\bibliographystyle{apalike}
{\small
\bibliography{References}}

\end{document}